\pdfoutput=1

\documentclass[11pt]{article}

\usepackage{acl}

\usepackage{times}
\usepackage{latexsym}

\usepackage[T1]{fontenc}

\usepackage[utf8]{inputenc}

\usepackage{microtype}

\usepackage{inconsolata}

\usepackage{amsmath}
\usepackage{times}
\usepackage{framed}
\usepackage{ragged2e}
\usepackage{bm}
\usepackage{soul}
\usepackage{latexsym}
\usepackage{subfigure}
\usepackage{amsthm}
\usepackage{graphicx}
\usepackage{float}
\usepackage{longtable}
\usepackage{booktabs} 
\usepackage{multirow}
\usepackage{multicol}
\usepackage{amssymb}
\usepackage{dashbox}%
\usepackage{multirow}
\usepackage{textcomp}
\usepackage{tcolorbox}
\usepackage{bbding}
\usepackage{fontawesome}
\usepackage{utfsym}
\usepackage{bbm}
\usepackage[ruled, vlined, linesnumbered]{algorithm2e}
\usepackage{mathtools}
\usepackage{adjustbox}
\usepackage[ruled,vlined]{algorithm2e}
\usepackage{color, soul}
\usepackage{pifont}
\usepackage{array}
\newcolumntype{P}[1]{>{\centering\arraybackslash}p{#1}}
\newcolumntype{M}[1]{>{\centering\arraybackslash}m{#1}}
\usepackage{cleveref}
\crefname{section}{§}{§§}
\Crefname{section}{§}{§§}
\crefname{figure}{Figure}{Figure}
\Crefname{figure}{Figure}{Figure}
\crefname{table}{Table}{Table}
\Crefname{table}{Table}{Table}
\newcommand\ourdataset{ELKEN\xspace}

\title{Event-level Knowledge Editing}

\author{Hao~Peng$^{1}$\thanks{\quad Equal contribution.}\hspace{0.0em}, Xiaozhi~Wang$^{1*}$,  Chunyang~Li$^{1}$, Kaisheng~Zeng$^{1}$, 
Jiangshan~Duo$^{1}$, \\ 
\textbf{Yixin~Cao$^{2}$, Lei~Hou$^{1}$, Juanzi Li$^{1}$} \\
$^1$Tsinghua University, Beijing, China\\
 $^2$Singapore Management University, Singapore\\
 \texttt{\{peng-h21, wangxz20\}@mails.tsinghua.edu.cn}
 }

\begin{document}
\maketitle
\begin{abstract}
Knowledge editing aims at updating knowledge of large language models (LLMs) to prevent them from becoming outdated. Existing work edits LLMs at the level of factual knowledge triplets. However, natural knowledge updates in the real world come from the occurrences of new events rather than direct changes in factual triplets. In this paper, we propose a new task setting: \textbf{event-level knowledge editing}, which directly edits new events into LLMs and improves over conventional triplet-level editing on (1) \textbf{Efficiency}. A single event edit leads to updates in multiple entailed knowledge triplets. (2) \textbf{Completeness}. Beyond updating factual knowledge, event-level editing also requires considering the event influences and updating LLMs' knowledge about future trends. We construct a high-quality event-level editing benchmark \ourdataset, consisting of $1,515$ event edits, $6,449$ questions about factual knowledge, and $10,150$ questions about future tendencies. We systematically evaluate the performance of various knowledge editing methods and LLMs on this benchmark. We find that \ourdataset poses significant challenges to existing knowledge editing approaches. Our codes and dataset are publicly released to facilitate further research.\footnote{\url{https://github.com/THU-KEG/Event-Level-Knowledge-Editing}}

\end{abstract}

\section{Introduction}
\begin{figure}
\centering
\includegraphics[width=1.0\linewidth]{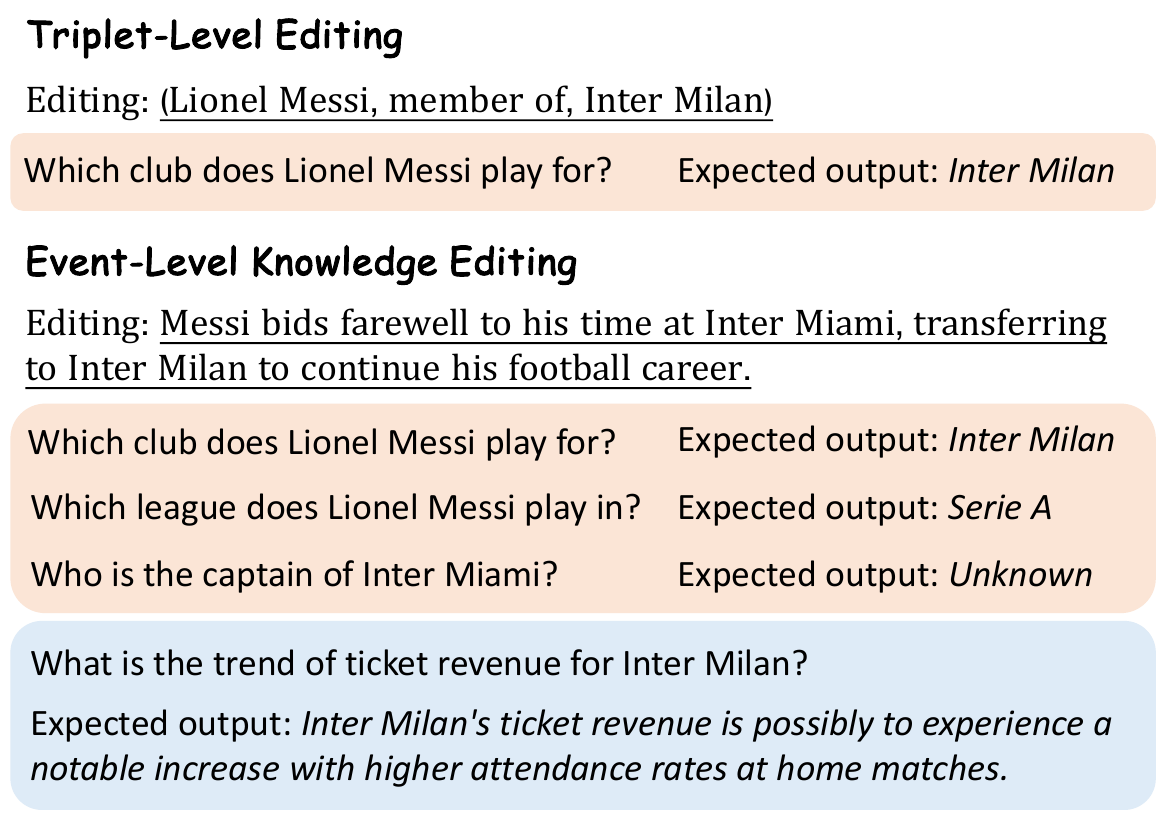}
    \caption{A counterfactual example for triplet-level and event-level knowledge editing. Triplet-level editing updates factual triplets into models. Event-level editing updates events into models, thus efficiently modifying \colorbox[cmyk]{0.03,0.09,0.11,0}{factual knowledge} and \colorbox[cmyk]{0.13,0.05,0.02,0}{tendencies} of models.}
    \label{fig:figure1}
\end{figure}

The world is constantly evolving, with new knowledge emerging frequently, leading to outdated or even misleading knowledge within language language models (LLMs). Therefore, numerous works focus on knowledge editing, aiming to update new knowledge into LLMs.~\citep{sinitsin2019editable, de2021editing, meng2022locating, meng2022mass, mitchell2022memory, wang2023knowledge, zheng2023can, zhang2024comprehensive}. 
Previous work defines knowledge editing as triplet-level editing, which edits factual knowledge triples into LLMs. As shown in Figure~\ref{fig:figure1}, supposing the triplet-level editing updates a new factual triplet \texttt{(Lionel Messi, member of, Inter Milan)} into LLMs, the model's answer to ``\textit{Which club does Lionel Messi play for?}'' should be changed to \texttt{Inter Milan}.

However, triplet-level editing is unnatural, as knowledge updates in the real world happen with new events rather than direct updates to knowledge triples. For example, in Figure~\ref{fig:figure1}, the update of the knowledge \texttt{(Lionel Messi, member of, Inter Milan)} is due to the event that Lionel Messi transfers to Inter Milan. Moreover, triplet-level editing has the following limitations: (1) \textbf{Inefficiency}. An event may update multiple factual triplets at once. As in Figure~\ref{fig:figure1}, Messi's transfer to Inter Milan updates several facts, including the sports club of Messi, the league where Messi plays, and Inter Miami's captain, etc. When a new event occurs, triplet-level editing needs to identify all affected triplets in advance before editing, which is time-consuming and labor-intensive. (2) \textbf{Incompleteness}. An event not only updates definite factual knowledge but can also affect potential tendencies of the future. For example, in Figure~\ref{fig:figure1}, Messi's transfer to Inter Milan could influence the tendency of ticket revenue for Inter Milan. Updating tendency knowledge in LLMs is crucial for enabling more reliable responses, such as event forecasting~\citep{zou2022forecasting, halawi2024approaching}.
However, existing triplet-level editing ignores the update in tendency knowledge.

Given the above issues, we propose a new task setting, \textbf{event-level knowledge editing}, aimed at editing newly occurred events into LLMs, thereby updating multiple factual knowledge and influenced tendencies at once. Event-level knowledge editing addresses the above limitations in two aspects: (1) \textbf{Updating all implicated facts at once}. 
Unlike triplet-level editing which requires explicitly identifying all the influenced triplets before editing, event-level editing aims at updating all the implicated factual triplets with a single event edit. For instance in Figure~\ref{fig:figure1}, after editing the event of Messi's transfer to Inter Milan into LLMs, the models should modify its multiple factual knowledge, such as the sports club of Messi, the league Messi plays in, and Messi's work location. This requires the model to infer all the factual triplets influenced by the event and also involves \textit{multi-hop reasoning}~\citep{zhong2023mquake}, such as the update of the league where Messi plays due to Messi playing for Inter Milan and Inter Milan being a club of the Serie A league.
Furthermore, we also consider the scenario of editing knowledge to \textit{unknown}~\citep{muresanu2024unlearnable}, which has not been explored to our knowledge. For example, in Figure~\ref{fig:figure1}, since Messi is no longer the captain of Inter Miami, and without additional information, Inter Miami's captain should be edited to \textit{unknown}. (2) \textbf{Updating tendency knowledge}. 
Beyond definite factual knowledge, event-level knowledge editing also enables updating the uncertain knowledge about future trends considering the new events. For example, in Figure~\ref{fig:figure1}, after editing the event of Messi's transfer to Inter Milan into LLMs, the models should adjust their knowledge on some tendencies, such as the tendency of ticket revenue for Inter Milan. This requires the model to understand the broad impact of event editing and possess \textit{common sense knowledge}~\citep{gupta2023editing}. For instance, in Figure~\ref{fig:figure1}, correctly predicting the tendency of ticket revenue for Inter Milan necessitates knowing that Messi is a football superstar and will draw more fans to watch Inter Milan's matches.

We construct a high-quality benchmark \ourdataset for event-level knowledge editing, including $1,515$ event edits along with $6,449$ questions for factual knowledge and $10,150$ questions for tendencies. 
To reduce costs and ensure that the construction methodology applies to other scenarios, we design a semi-automatic construction process. For factual knowledge, we manually create several event templates and their impacted triplets. We sample entities from Wikidata~\citep{vrandevcic2014wikidata} to instantiate the templates and obtain event edits and question-answer pairs. We then use GPT-3.5~\citep{chatgpt} to paraphrase the event edits to get the final diverse edits. For tendencies, we first reuse event edits generated for factual knowledge and augment them with events having a broader impact. We use GPT-3.5 to generate tendency-related question-answer pairs and verify the generated data with human annotation.

We conduct systematic experiments and analysis on \ourdataset, evaluating $5$ representative methods, including Fine-tuning~\citep{yao2023editing}, Spare and Dense Retrieval~\citep{akyurek2023dune}, SERAC~\citep{mitchell2022memory}, and In-Context Editing (ICE)~\citep{akyurek2023dune}, and $6$ language models, including GPT-J~\citep{gpt-j}, TULU 2~\citep{ivison2023camels}, Mistral 7B~\citep{jiang2023mistral}, GPT-3.5~\citep{chatgpt}, GPT-4~\citep{openai2023gpt}, and Gemini Pro~\citep{team2023gemini}. We find that the event-level knowledge editing task presents significant challenges to existing editing methods and models, which highlights the importance of future research.

\section{Event-level Knowledge Editing}
\begin{figure*}[t]
    \centering
    \includegraphics[width=1.00\linewidth]{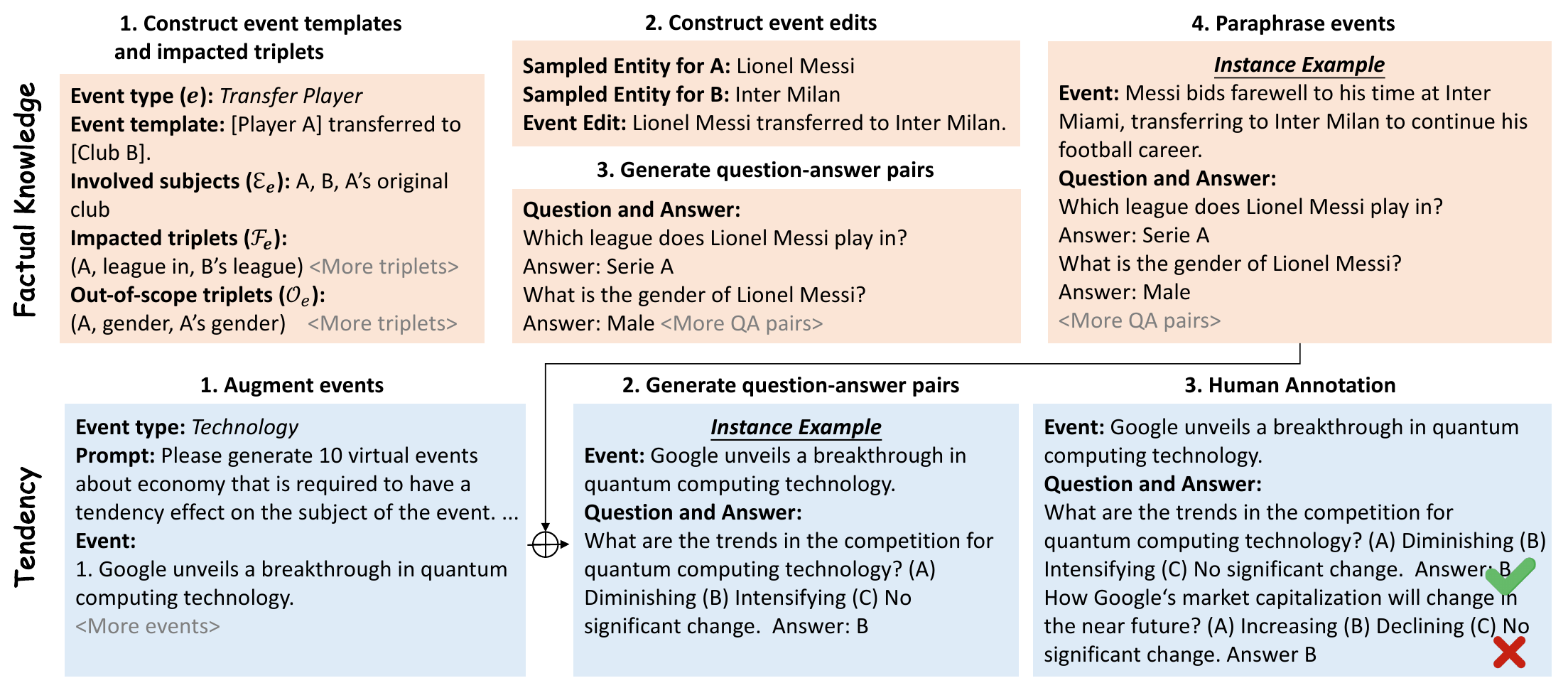}
    \caption{The overall construction process of \ourdataset, including two categories of question-answer pairs: \texttt{Factual Knowledge} and \texttt{Tendency}. \underline{\textit{Instance Example}} demonstrates a sample of the data.}
    \label{fig:bench_construct}
\end{figure*}

\subsection{Task Definition}
\label{sec:task_form}

Event-level knowledge editing aims to edit events into LLMs, thereby updating both influenced definite factual knowledge and uncertain knowledge about future tendencies at once. The objectives and challenges of event-level knowledge editing primarily include two aspects: (1) \textbf{Updating all implicated facts at once}. An event edit can update multiple factual knowledge at once, and determining its scope is challenging. Additionally, updating corresponding factual knowledge about an event edit may involve \textit{multi-hop reasoning}~\citep{zhong2023mquake} and editing knowledge to \textit{unknown}~\citep{muresanu2024unlearnable}. (2) \textbf{Updating tendency knowledge}. An event edit can also update uncertain knowledge about future tendencies, and identifying the broad tendency impacts of an event edit is challenging, usually requiring common sense knowledge~\citep{gupta2023editing}.

Formally, given an event edit $e$, $f_{\theta}$ represents the model before the edit, with $\theta$ denoting the model's parameters, and $f_{\theta_{e}}$ denotes the model after editing the edit $e$. $\mathcal{F}_e$ and $\mathcal{T}_e$ represent the scope of factual knowledge and tendency impacted by $e$, respectively. We refer to the questions in $\mathcal{F}_e \cup \mathcal{T}_e$ as \textbf{in-scope} questions. Moreover, the editing process should not affect the model's unrelated knowledge~\citep{yao2023editing}, which are referred to as \textbf{out-of-scope} knowledge and denoted as $\mathcal{O}_e$. The goal of event-level knowledge editing is as follows:

\begin{equation}
f_{\theta_{e}}(x) = \left\{
\begin{array}{lcl}
   y_e  & & x \in  \mathcal{F}_e \cup \mathcal{T}_e\\
   f_{\theta}(x) & &  x \in  \mathcal{O}_e
\end{array}
\right.
\end{equation}
$y_e$ is the expected answer after editing.
Based on this objective of event-level knowledge editing, we assess the editing methods from two dimensions: reliability and locality.

\textbf{Reliability} assesses whether the edited model answers as expected, evaluating the accuracy of answers to in-scope questions about $\mathcal{F}_e \cup \mathcal{T}_e$: 

\begin{equation}
    \mathbb{E}_{(x, y_e) \in \mathcal{F}_e \cup \mathcal{T}_e}\mathbbm{1}\{\text{argmax}_yf_{\theta_e}(y|x)=y_e\}.
\end{equation}

\textbf{Locality} means that the editing should not affect the model's answers to unrelated questions, evaluating the consistency of the model's answers to the unrelated questions in $\mathcal{O}_e$ before and after editing:

\begin{equation}
    \mathbb{E}_{(x, y_e) \in \mathcal{O}_e} \mathbbm{1}\{f_{\theta_{e}}(y|x)=f_\theta(y|x)\}.
\end{equation}

\subsection{Benchmark Construction}
\label{sec:bench_construct}
Our ELKEN benchmark consists of data for factual knowledge impacts (\texttt{Factual Knowledge}) and tendency impacts (\texttt{Tendency}). \cref{fig:bench_construct} illustrates the overall data construction process, and \cref{tab:statistics} shows the data statistics. More construction details and comparisons to existing triplet-level editing datasets are shown in \cref{sec:appendix_data}.

\paragraph{Construction of \texttt{Factual Knowledge}}
Unlike the data construction of triplet-level editing, which only requires replacing entities within triplets for constructing edits and question-answer pairs~\citep{yao2023editing}, the construction of event-level editing is more complex, as identifying the impact scope of an event is difficult. 

To this end, we propose a semi-automatic approach that conserves human efforts while ensuring data quality and is transferable to other scenarios. As illustrated in Figure~\ref{fig:bench_construct}, the overall construction process of \texttt{Factual Knowledge} consists of $4$ steps. (1) \textbf{Constructing event templates and their impacted triplets}. 
Our method for determining the impact scope is similar to ``Ripple Effects''~\citep{cohen2023evaluating}, involving manual efforts, but the impact scope of events is broader, involving more subjects and triplets.
We first select $16$ common event types from MAVEN~\citep{wang2020maven, wang2023maven} and ACE 2005~\citep{walker2006ace} that are likely to lead to changes in factual knowledge. We also select $81$ widely-used relationships from Wikidata, denoted as $\mathcal{R}$. For each event type, we manually construct an event template. For example, for the ``\textit{Transfer Player}'' type, its template is ``\textit{[Player A] transferred to [Club B]}'' with placeholders \texttt{A} and \texttt{B}. We then manually identify the directly involved subjects of the event $e$, denoted as $\mathcal{E}_e$. For each subject $s$ in $\mathcal{E}_e$, we refer to $\mathcal{R}$ and manually identify the scope of triplets $\mathcal{F}_s$ of $s$ impacted by this event, which consists of ($s$, $r$, $o*$), where $r \in \mathcal{R}$ and $o*$ denotes the updated answer. 
For instance in Figure~\ref{fig:bench_construct}, $\mathcal{E}_e$ includes \texttt{A}, \texttt{B}, and \texttt{A}'s original club. $\mathcal{F}_s$ of the subject A includes \texttt{(A, member of, B)}, etc. We then aggregate $\mathcal{F}_s$ of each subject in $\mathcal{E}_e$ as $\mathcal{F}_e$. We adpot the same method to construct out-of-scope triplets $\mathcal{O}_e$, which consists of ($s$, $r$, $o$) where $s \in \mathcal{E}_e$, $r \in \mathcal{R} \setminus \{r | (s, r, o*) \in \mathcal{F}_e\}$, and $o$ denotes the ground truth answer from Wikidata.  
To ensure a comprehensive and accurate identification of an event's impact scope, we involve $3$ annotators to identify the impact scope, $\mathcal{F}_e$, and then we assemble all their annotations. 
(2) \textbf{Constructing event edits}. 
We instantiate event templates to create event edits. Specifically, we construct edits by sampling the $100$ most frequent entities of corresponding types from Wikidata~\citep{vrandevcic2014wikidata} based on the frequency counts from Wikipedia\footnote{\url{https://en.wikipedia.org}} to replace placeholders with specific entities. 
(3) \textbf{Generating question-answer pairs}. 
With the instantiated event edits and impacted triplets, we generate in-scope question-answer pairs. For each triplet ($s$, $r$, $o*$) in $\mathcal{F}_e$,  we adopt predefined rules to transform ($s$, $r$) as a question and take the instantiated $o*$ as the answer. For each triplet in $\mathcal{O}_e$, we adopt the same method to construct out-of-scope question-answer pairs.
(4) \textbf{Paraphrasing event edits}.
Finally, to make the expressions of event edits more natural and enhance linguistic diversity, we use LLMs to paraphrase the instantiated event templates and generate the final event edits. Specifically, we employ GPT-3.5~\citep{chatgpt} in paraphrasing as GPT-3.5 is demonstrated as an effective paraphraser~\citep{cegin2023chatgpt}. We manually review and verify each paraphrased event and find little noise.
Finally, we obtain $841$ event edits, $3,307$ in-scope questions, and $3,142$ out-of-scope questions in \texttt{Factual Knowledge}. 
We divide the data into a training set and a test set by event types. 
\begin{table}[t]
    \centering
    \small
    \resizebox{\columnwidth}{!}{
    \begin{tabular}{llrr}
     \toprule
     & & Train & Test \\
     \midrule
     & \#Event Edits & $671$ & $844$\\
     \midrule
     \multirow{2}{*}{\texttt{Factual Knowledge}} & \#In-scope Q & $971$ & $2,171$ \\
     & \#Out-of-scope Q & $1,325$ & $1,982$ \\
     \midrule
     \multirow{2}{*}{\texttt{Tendency}} 
     & \#In-scope Q  & $3,889$ & $3,968$ \\
     & \#Out-of-scope Q & $1,353$ & $940$ \\
    \bottomrule
    \end{tabular}
    }
    \caption{Overall statistics of \ourdataset. Q: Question. 
    \ourdataset comprises two categories of question-answer pairs: \texttt{Factual Knowledge} and \texttt{Tendency}.
    \texttt{Tendency} has two evaluation formats: \texttt{Tendency-M} for multiple choice and \texttt{Tendency-G} for open-ended generation.
    }
    \label{tab:statistics}
\end{table}

\paragraph{Construction of \texttt{Tendency}}
\texttt{Tendency} reuses event edits from \texttt{Factual Knowledge}, which are usually specific and have limited broad impacts. To comprehensively evaluate LLMs' understanding of the broader tendency impact of events, we augment some event edits for \texttt{Tendency} by generating new events with LLMs. To identify the scope of impacted tendencies ($\mathcal{T}_e$), we preliminarily examine with human annotators and find manually crafted question-answer pairs about tendencies homogeneous and the process is labor-intensive. Previous work has shown that LLMs can provide reasonable predictions about future tendencies~\citep{wang2023maven,halawi2024approaching}. Therefore, we adopt LLMs first to generate a rich set of question-answer pairs about tendencies, followed by manual annotation.
Specifically, as shown in Figure~\ref{fig:bench_construct}, the construction process of \texttt{Tendency} includes $3$ steps: 
(1) \textbf{Augmenting events}. We collect $18$ event topics, such as \textit{politics}, \textit{sports}, etc., and use GPT-3.5~\citep{chatgpt} to generate $100$ counterfactual events for each event topic. We then filter out repeated events following \citet{wang2022self}. 
(2) \textbf{Generating question-answer pairs}. We use GPT-3.5 to generate tendency-related questions and answers.
Although exhausting all possible tendencies is impossible, we prompt GPT-3.5 to generate rich and representative question-answer pairs through instructions and diverse demonstrations.
For each event edit, we generate $6$ in-scope and $2$ out-of-scope question-answer pairs, each consisting of one question, three choices, and one answer. We manually assess $100$ sampled questions and answers and find the accuracy rate is about $85\%$ and the questions exhibit great diversity (covering much more topics than human-written questions), indicating the high quality of the model-generated data. 
(3) \textbf{Human Annotation}. Same as \texttt{Factual Knowledge}, we divide the data into a training set and a test set by event topics and types. 
To ensure the benchmark's quality, we manually annotate the whole test set to verify the model-generated questions and answers. 
To maintain annotation quality, all data are annotated twice and similar questions are filtered out.
The final inter-annotator agreement reaches $95.6\%$. We filter out questions with inconsistent annotations and those whose answers are marked as incorrect. Finally, we obtain $1,515$ event edits, including $841$ edits from \texttt{Factual Knowledge} and $674$ newly generated edits, $7,857$ in-scope questions, and $2,293$ out-of-scope questions.

\section{Experiments}

\begin{table*}[t]
    \centering
    \small
    \begin{adjustbox}{max width=1\linewidth}
{
    \begin{tabular}{ll|rrr|rrr|rrr|r}
    \toprule
    \multirow{2}{*}{Model} &\multirow{2}{*}{Method} & \multicolumn{3}{c|}{\texttt{Factual Knowledge}} &
    \multicolumn{3}{c|}{\texttt{Tendency-M}} & 
    \multicolumn{3}{c|}{\texttt{Tendency-G}} & \multirow{2}{*}{E-Level} \\
    & & E-Level & Q-Level & Locality & E-Level & Q-Level & Locality  & E-Level & Q-Level & Locality & \\
    \midrule
    \multirow{5}{*}{GPT-J} & Fine-tuning & $0.0$&$2.3$&$67.2$ &$4.8$&$41.3$&$99.3$ & $0.2$&$4.8$ & $87.7$ & $0.1$\\
    & Sparse Retrieval & $15.1$&$44.2$&$29.4$& $7.5$&$49.3$&$42.8$ &$0.7$&$12.0$ & $51.6$ & $0.2$\\
    & Dense Retrieval & $16.7$&$46.7$&$28.8$ &$7.3$&$49.2$&$40.6$ & $0.6$&$12.5$ & $53.8$ & $0.1$ \\
    & SERAC & $4.3$&$20.4$&$65.1$ & $8.2$&$49.4$&$74.4$ & $0.0$&$4.2$ & $80.3$ & $0.0$\\
    & ICE &$17.1$&$49.9$&$29.1$ & $7.5$&$49.6$&$41.6$ & $1.6$&$13.7$ & $54.3$ & $0.9$\\
    \midrule
    \multirow{5}{*}{TULU 2} &Fine-tuning & $0.0$&$4.7$&$\mathbf{90.6}$& $10.3$&$53.9$&$\mathbf{100.0}$ &$5.1$&$36.3$&$81.3$ & $2.9$\\
    & Sparse Retrieval & $26.4$&$56.3$&$53.7$ & $16.4$&$63.4$&$40.3$ & $2.9$&$34.0$ & $28.5$ & $2.6$\\
    & Dense Retrieval & $28.7$&$59.3$&$52.8$ & $24.9$&$69.5$&$42.1$ & $7.1$&$41.4$ & $25.2$ & $5.5$\\
    & SERAC & $7.0$&$30.4$&$89.1$ & $18.9$&$65.3$&$59.0$ & $4.6$&$38.0$ & $76.7$ & $2.9$\\
    & ICE & $30.5$&$63.8$&$53.7$ & $34.1$&$75.9$&$39.3$ &$9.5$&$43.8$& $25.1$ & $8.6$\\
    \midrule
    \multirow{5}{*}{Mistral 7B} &Fine-tuning & $0.2$&$4.1$&$66.8$ & $21.5$&$65.4$&$\mathbf{100.0}$ & $19.1$&$59.7$ & $77.6$ & $10.5$\\
    & Sparse Retrieval & $24.1$&$57.5$&$39.6$ & $28.0$&$72.5$&$34.7$ &$6.1$&$43.9$ & $28.6$ & $3.9$\\
    & Dense Retrieval & $25.6$&$60.4$&$39.1$ & $40.5$&$79.2$&$37.6$ & $12.8$&$53.6$ & $21.4$ & $9.9$\\
    & SERAC & $7.4$&$27.4$&$71.1$ & $37.4$&$76.4$&$59.8$ &$14.1$&$54.7$ & $74.7$ & $8.6$\\
    & ICE & $26.6$&$64.5$&$39.8$ & $60.1$&$88.0$&$35.6$ &$21.5$&$59.6$ & $22.8$ & $16.7$\\
    \midrule
    \multirow{4}{*}{GPT-3.5} &
    Sparse Retrieval & $16.9$&$55.0$&$33.0$& $49.4$&$82.4$&$41.2$ & $10.4$&$48.7$ & $23.9$ & $7.2$\\
    & Dense Retrieval & $18.4$&$60.2$&$30.6$ & $57.6$&$86.0$&$46.6$ & $21.5$&$60.1$ & $19.3$ & $16.3$\\
    & SERAC & $5.2$&$27.1$&$71.2$& $56.0$&$84.9$&$70.3$ & $17.9$&$57.7$ & $70.6$ & $11.7$\\
    & ICE & $20.0$&$63.1$&$32.7$ & $\mathbf{71.6}$&$\mathbf{91.6}$&$41.9$ & $33.8$&$66.6$ & $20.1$ & $27.1$\\
    \midrule
    \multirow{4}{*}{GPT-4} &Sparse Retrieval & $34.2$&$64.7$&$56.6$ & $30.6$&$71.8$&$52.0$ & $14.5$&$58.4$ & $34.0$ & $9.8$\\
    & Dense Retrieval & $36.5$&$68.7$&$56.1$ &$46.5$&$80.6$&$51.8$ & $24.3$&$66.1$ & $31.0$ & $18.9$\\
    & SERAC &$9.7$&$31.4$&$80.6$ & $45.8$&$81.4$&$92.3$ & $26.5$&$65.7$ & $\mathbf{93.1}$ & $15.2$\\
    & ICE & $\mathbf{39.0}$&$\mathbf{73.5}$&$56.9$ & $66.4$&$89.3$&$49.8$ & $\mathbf{40.3}$&$\mathbf{73.0}$ & $31.9$ & $\mathbf{29.2}$ \\
    \midrule
    \multirow{4}{*}{Gemini Pro} &Sparse Retrieval & $24.3$&$60.3$&$30.3$ & $13.8$&$57.3$&$38.0$ &  $2.8$&$29.8$ & $33.7$ & $2.4$\\
    & Dense Retrieval & $25.2$&$63.7$&$30.5$ & $28.2$&$67.2$&$43.4$& $6.6$&$39.5$ & $33.9$& $5.6$\\
    & SERAC & $6.0$&$28.3$&$72.1$ & $31.1$&$72.3$&$77.0$ & $8.4$&$45.7$& $70.3$ & $5.5$\\
    & ICE & $24.3$&$65.6$&$41.6$ &$41.9$&$75.2$&$40.6$ &$7.4$&$38.1$& $39.7$ & $7.2$\\
    \midrule

    \end{tabular}
}
\end{adjustbox}
    \caption{Experimental results (\%) of all investigated methods and models on \ourdataset. E-Level: Edit-level reliability. Q-Level: Question-level reliability. The results on \texttt{Tendency-G} are the percentages of full-mark of \textit{overall} scores. The rightmost column, E-Level, displays the overall reliability considering $\mathcal{F}_e \cup \mathcal{T}_e$.}
    \label{tab:main_results}
\end{table*}

\subsection{Experimental Setup}
\label{sec:setup}
As mentioned in \cref{sec:task_form}, the evaluation metrics have two dimensions: reliability and locality.
For \texttt{Factual Knowledge}, the reliability is accuracy and the locality is the proportion of answers that are the same before and after editing. These metrics are calculated using \textbf{exact match}.
For \texttt{Tendency}, the evaluation is more complicated since the tendency judgment is an open-ended generation task by nature, and we evaluate it under two settings. For the first generation setting (\texttt{Tendency-G}), we adopt an automated evaluation method using GPT-4~\citep{openai2023gpt} as the evaluator, which has been verified as an effective~\citep{bai2023benchmarking, li2024leveraging}. Specifically, for the evaluation of reliability, we use the correct option of each question as the reference, comprehensively scoring the editing methods in $3$ dimensions: \textit{correctness}, \textit{coherence}, and \textit{comprehensiveness}. We also ask GPT-4 to give an \textit{overall} score. Similar to the previous scoring-based evaluation method~\citep{li2024leveraging}, all scores are integers scaling from $1$ to $5$, with $5$ being the best.
For the evaluation of locality under \texttt{Tendency-G}, we utilize GPT-4 to assess the consistency of the model's responses to out-of-scope questions in $\mathcal{O}_e$ before and after editing, also using an integer score from $1$ to $5$, with $5$ being the most similar. 
To avoid the potential risk of GPT-4 evaluation and provide an alternative metric for \texttt{Tendency}, we also adopt a \textbf{multiple-choice} evaluation setting (\texttt{Tendency-M}), which is the same with \texttt{Factual Knowledge}, i.e., using extract match to calculate the reliability and locality. Experimental details of using GPT-4 scorer on \texttt{Tendency-G} are shown in \cref{sec:app_gpt4}.

For reliability, we employ evaluations at two levels: \textit{question-level} and \textit{edit-level}. The \textit{question-level} evaluation assesses the reliability of each individual question. For \texttt{Tendency-G} scores evaluated using GPT-4, similar to \citet{bai2023benchmarking}, we present the percentages of responses with full marks, \textit{i.e.}, scored $5$ points. The \textit{edit-level} evaluation assesses the reliability of each edit. An edit is reliable only if all questions in $\mathcal{F}_e$ (for \texttt{Factual Knowledge}) or $\mathcal{T}_e$ (for \texttt{Tendency-M} and \texttt{Tendency-G}) are answered correctly \textit{or} the \textit{overall} scores of answers are all full-mark.

\subsection{Investigated Editing Methods and Models}
\label{sec:methods}
We evaluate various advanced editing methods on our benchmark, including: (1) \textbf{Fine-tuning}~\citep{zhu2020modifying, meng2022mass, akyurek2023dune}. Fine-tuning is a vanilla method of editing, involving direct learning the new edits by fine-tuning model parameters. In our experiments, we fine-tune all the parameters of models on edits in the test set using a language modeling objective. However, this method has high computational costs and may also lead to catastrophic forgetting~\citep{luo2023empirical}. (2) \textbf{Retrieval}~\citep{madaan-etal-2022-memory, zhong2023mquake}. This method is memory-based, which stores all edits in an external memory. When posed with a question, this method first retrieves the most matching edit to use as context along with the question for input into the models. In our experiments, we used BM25 and E5~\citep{wang2022text} as the retrieval methods, named \textit{sparse retrieval} and \textit{dense retrieval}, respectively. (3) \textbf{SERAC}~\citep{mitchell2022memory}. SERAC is also a memory-based method. This approach trains a scope classifier to determine whether a question requires retrieving a corresponding edit for answering. If retrieval is necessary, the retrieved edit and the question are input together into a \textit{counterfactual} model for answering; otherwise, the question alone is input into the vanilla pre-trained model. In our implementation, we train a cross-encoder classifier~\citep{mitchell2022memory} based on ELECTRA~\citep{clark2019electra} and we use the same pre-trained model as the \textit{counterfactual} model, which is the same as in \citet{akyurek2023dune}.
(4) \textbf{ICE}. This method takes the ground truth edit as context along with the question as input into pre-trained models, directly evaluating whether the model can understand the scope of the edit and correctly answer the corresponding questions. 
We do not evaluate some advanced approaches, such as the Locate-Then-Edit methods~\citep{dai-etal-2022-knowledge, meng2022locating, meng2022mass, li2023pmet}, because these approaches are specifically designed for triple-level editing and are not applicable to event-level knowledge editing.

We adopt several advanced language models as the base models to implement the aforementioned methods. We employ three open-source models, including GPT-J~\citep{gpt-j}, TULU 2~\citep{ivison2023camels}, Mistral 7B~\citep{jiang2023mistral}, and three powerful proprietary models, including GPT-3.5~\citep{chatgpt}, GPT-4~\citep{openai2023gpt}, and Gemini Pro~\citep{team2023gemini}. The implementation details of the editing method and automated evaluation are in \cref{sec:appendix_exp}.

\subsection{Experimental Results}
\label{sec:results}
The experimental results are shown in Table~\ref{tab:main_results}, and we have the following general observations:
(1) Existing methods exhibit moderate performance on \ourdataset. Even the best-performing method (ICE + GPT-4) falls short, which indicates the significant challenge posed by the new event-level knowledge editing setting.
(2) The question-level reliability scores on \ourdataset are much lower than those in triplet-level editing. For instance, SERAC can achieve nearly $100\%$ reliability~\citep{yao2023editing} in triplet-level editing.
Moreover, the reliability scores of event-level evaluations are further lower than those of question-level evaluations.
This suggests that recognizing the impact scope of event editing is a novel challenge of our task. The impact scope of triplet-level editing typically confines to edited triplets themselves, while that of event-level knowledge editing extends to multiple factual and tendency knowledge.
(3) The locality scores on \ourdataset are also generally lower than those in triplet-level editing. For example, SERAC + GPT-J achieves nearly $100\%$ locality in triplet-level editing~\citep{yao2023editing} but only attains about $80\%$ and $65.1\%$ in \texttt{Tendency} and \texttt{Factual Knowledge}, respectively. This may be due to the broad impact range of event edits, making the models struggle to ensure the locality of edits, which poses new challenges to existing methods.
(4) On in-scope questions of \texttt{Tendency-G}, the full-mark rate is lower compared to reliability scores on \texttt{Tendency-M}. This is because \texttt{Tendency-G} not only assesses the tendency correctness of answers but also evaluates the coherence and comprehensiveness. This indicates that although the model may correctly identify the tendency of a question, it struggles to provide comprehensive and reasonable explanations.

\section{Further Analysis}
This section presents some further analyses. Unless otherwise specified, the experimental results are from the ICE method, with \textit{question level} reliability scores. More results are placed in \cref{sec:appendix_result}.

\subsection{Analysis on \textit{Unknown} Questions}
\begin{table}[t]
    \centering
    \small
    \begin{tabular}{lrrr}
    \toprule
    Model & \textit{Unknown} & \textit{Known} & Overall \\
    \midrule
    GPT-J & $28.2$ & $63.9$ & $49.9$ \\
    TULU 2 &$44.2$ & $76.5$ & $63.8$ \\
    Mistral 7B &$50.1$ & $73.8$ & $64.5$ \\
    GPT-3.5 &$\mathbf{67.8}$ & $60.0$ & $63.1$ \\
    GPT-4 &$63.6$ & $\mathbf{79.9}$ & $\mathbf{73.5}$ \\
    Gemini Pro &$54.0$ & $73.2$ & $65.6$ \\
    \bottomrule
    \end{tabular}
    \caption{Reliability (\%) on \textit{Unknown} and \textit{Known} questions of \texttt{Factual Knowledge} in \ourdataset.}
    \label{tab:unknown}
\end{table}
As mentioned in \cref{sec:bench_construct}, the editing process may render some facts as unknown, such as \texttt{Inter Miami's captain} in Figure~\ref{fig:figure1}. 
This process is a form of knowledge deletion or unlearning~\citep{si2023knowledge}, which has not been covered by previous editing work. We further investigate whether the LLMs recognize that certain knowledge should be deleted based on edits, namely answering ``\textit{unknown}'' to relevant queries.
Specifically, in \ourdataset, there are $797$ in-scope questions with answers marked as \textit{Unknown} and the remaining $1,374$ in-scope questions with \textit{Known} answers being specific entities. We observe the model's performance on these different data types, with results presented in Table~\ref{tab:unknown}. 

Our observations are as follows: (1) In general, models exhibit significantly lower reliability on \textit{Unknown} questions compared to \textit{Known} questions, except for GPT-3.5. This suggests that deleting corresponding outdated knowledge based on edits remains a challenge for current methods. (2) GPT-J performs notably worse on \textit{Unknown} questions than other aligned models, indicating that alignment, e.g., instruction-tuning~\citep{wei2021finetuned, chung2022scaling} or RLHF~\citep{ouyang2022training}, can enhance the models' ability to delete knowledge through human instructions. The ``editing to \textit{Unknown}'' questions included in \ourdataset presents new challenges for existing knowledge editing methods and necessitates further efforts, such as incorporating knowledge unlearning methods~\citep{si2023knowledge, muresanu2024unlearnable}.

\subsection{Analysis on Questions needing Background Knowledge}
\begin{table}[t]
    \centering
    \small
    \begin{tabular}{l|rr|r}
    \toprule
    Model & K. Needed & No K. Needed & Recall \\
    \midrule
    GPT-J &$44.0$ & $55.6$ & $43.5$ \\
    TULU 2 &$59.9$ & $65.9$ & $73.3$ \\
    Mistral 7B &$49.8$ & $69.1$ & $61.3$ \\
    GPT-3.5 &$23.3$ & $76.0$ & $82.2$ \\
    GPT-4 &$\mathbf{62.0}$ & $\mathbf{79.5}$ & $\mathbf{95.7}$ \\
    Gemini Pro &$52.7$ & $70.5$ & $84.5$ \\
    \bottomrule
    \end{tabular}
    \caption{Reliability (\%) on questions needing background knowledge (\textit{K. Needed}) versus questions not requiring background knowledge (\textit{No K. Needed}) and recall rate (\%) of background knowledge needed.}
    \label{tab:require_know}
\end{table}

As noticed by \citet{zhong2023mquake}, LLMs may require background knowledge to answer certain questions. There are also such questions in our benchmark \ourdataset.
For instance, in Figure~\ref{fig:figure1}, correctly answering the question ``\textit{Which league does Lionel Messi play in?}'' necessitates the knowledge of ``\textit{Inter Milan is a club of the Serie A league}''. Correctly answering these questions also involves \textit{multi-hop reasoning}, as the update of the league where Lionel Messi plays due to Messi playing for Inter Milan and Inter Milan being a club of the Serie A league.
Therefore, successfully editing models not only requires the model to understand the editing scope of the edit, which requires \textit{multi-hop reasoning} abilities, but also relies on the model's background knowledge. In \ourdataset, there are $393$ questions that need background knowledge for answers, which are marked during the construction of \ourdataset. We observe the model's performance on the questions and find that the performance on questions requiring background knowledge is significantly lower, as shown in Table~\ref{tab:require_know}.

We further analyze the reasons for the lower performance on questions requiring background knowledge. We assess the model's recall rate for the knowledge required to answer questions, with results presented in Table~\ref{tab:require_know}. We find that most models could recall a substantial proportion of the knowledge. However, their accuracy on the corresponding questions is much lower, indicating that the main reason for errors in these cases is the model's failure to recognize the editing scope requiring \textit{multi-hop reasoning}, which poses a significant challenge to existing methods.

\subsection{Comprehensive Evaluation on \texttt{Tendency-G} of \ourdataset}
\begin{table}[t]
    \centering
    \small
        \begin{adjustbox}{max width=1\linewidth}
{
    \begin{tabular}{lrrr}
    \toprule
    Model & Correctness & Coherence & Comprehensiveness \\ 
    \midrule
    GPT-J & $41.5$ & $11.8$ & $4.7$ \\
    TULU 2 & $55.4$ & $40.7$ & $14.2$ \\
    Mistral 7B & $62.3$ & $58.9$ & $26.2$ \\
    GPT-3.5 &$69.8$ & $67.4$ & $22.7$ \\
    GPT-4 &$\mathbf{71.7}$ & $\mathbf{82.1}$ & $\mathbf{76.8}$ \\
    Gemini Pro &$38.9$ & $42.1$ & $38.8$ \\
    \bottomrule
    \end{tabular}
    }
    \end{adjustbox}
    \caption{Full-mark rate results (\%) across three dimensions on \texttt{Tendency-G} of \ourdataset.}
    \label{tab:comprehensive}
\end{table}

As mentioned in \cref{sec:setup}, we conduct a systematic evaluation across $3$ dimensions on \texttt{Tendency-G} of \ourdataset. We present the results of this systematic evaluation in Table~\ref{tab:comprehensive}. We find that: (1) For correctness, the results evaluated by GPT-4 and those on \texttt{Tendency-M} are roughly similar in the model's relative performance\footnote{
The significant discrepancy in Gemini Pro's performance between \texttt{Tendency-G} and \texttt{Tendency-M} is primarily due to Gemini Pro often being unable to respond on \texttt{Tendency-G} due to triggering safety concerns.}, but the results on \texttt{Tendency-G} are significantly lower. One reason is that the evaluation here employs a full-mark scheme, which is more stringent. If we consider results with correctness $\geq 4$ as correct, then the gap between \texttt{Tendency-G} and \texttt{Tendency-M} scores is generally within $10\%$. 
Therefore, if one worries about the GPT-4 evaluation quality, one can always refer to the \texttt{Tendency-M} results. (2) Some models, e.g., GPT-3.5, despite high correctness, score low on coherence or comprehensiveness, indicating that while the model could correctly answer the tendencies of the questions, it fails to provide reasonable or comprehensive explanations, which is also undesirable. 
This suggests that a comprehensive evaluation across multiple dimensions is necessary.

\subsection{Human Evaluation of GPT-4 Scorer}
To validate the effectiveness of using GPT-4 as a scorer in the \texttt{Tendency-G} evaluation, we conduct a manual review of GPT-4's scoring. Specifically, we randomly sample $120$ questions and corresponding model-generated answers, with $60$ from Mistral 7B and $60$ from GPT-4. One of our authors scores this data. Similar to previous work~\citep{bai2023benchmarking, chan2023chateval}, we calculate Spearman's $\rho$ and Kendall's $\tau$ coefficients between the model's \textit{overall} scores and the manually assigned \textit{overall} scores, which are $74.4\%$ and $69.8\%$, respectively. These results indicate a strong positive correlation between scores given by GPT-4 and humans. This suggests that GPT-4's scoring generally aligns with human assessment but still leaves room for improvement. Additionally, GPT-4 tends to overestimate LLMs' performance, with an average score of $4.34$ compared to the human-assigned average of $4.15$. Nonetheless, as an automated, low-cost evaluation approach, it is sufficiently effective.

\section{Related Work}

\paragraph{Knowledge Editing Datasets.}
Most existing knowledge editing datasets assess triplet-level editing, including ZsRE~\citep{levy2017zero}, CounterFact~\citep{meng2022locating}, Fact Verification~\citep{mitchell2022memory},
Calibration~\citep{dong2022calibrating}, MQuAKE~\citep{zhong2023mquake}, RaKE~\citep{wei2023assessing}, RIPPLEEDITS~\citep{cohen2023evaluating}, etc. 
Some datasets evaluate various editing settings, such as \citet{mitchell2021fast} incorporating a piece of scrambled text into the model; \citet{mitchell2022memory} editing the sentiment on a specific topic into the model; \citet{wu2023eva} editing triplets into LLMs by inputting raw documents; \citet{akyurek2023dune} introducing a unified editing task, defining edits as any arbitrary natural language. Our benchmark evaluates event-level knowledge editing, a form that enables efficient and comprehensive updating of knowledge within the model. 

\paragraph{Knowledge Editing Methods.}
Previous knowledge editing methods primarily focus on triplet-level editing, encompassing the following categories: (1) \textit{Memory-based method}~\citep{mitchell2022memory, madaan2022memory, zhong2023mquake, zheng2023can}. This approach stores edits in an external memory, then uses a retriever to retrieve the most relevant edit as context for question answering. Typically, the base model does not require additional parameter updating.
(2) \textit{Locate-Then-Edit method}~\citep{dai-etal-2022-knowledge, meng2022locating, meng2022mass, li2023pmet, ma2023untying, hase2024does, gupta2024rebuilding}. This approach initially identifies the specific location of the knowledge to be edited within the base model, usually a neuron, and then modifies this neuron to significantly reduce the impact of the edit on other knowledge, making it a promising approach to knowledge editing. (3) \textit{Hyper-network method}~\citep{de2021editing, mitchell2021fast, tan2023massive}. This method generally employs an additional neural network to learn from edits, generating corresponding parameter offsets for the base model to incorporate the knowledge edits. 
The above-mentioned \textit{Locate-Then-Edit} and \textit{Hyper-network} methods are typically designed specifically for triplet-level editing, involving entities or relations, and thus cannot be straightforwardly applied to event-level editing. In this work, we mainly evaluate \textit{memory-based method} and in-context editing. We leave the development of advanced editing methods for event-level knowledge editing as the future work.

\section{Conclusion}
In this paper, we introduce event-level knowledge editing, aimed at editing newly occurred events into LLMs to update multiple factual knowledge and influenced tendencies at once. We propose a semi-automated data construction approach and create a high-quality benchmark \ourdataset. We conduct extensive experiments and find that existing methods struggle to delineate the scope of event edits, which poses significant challenges.

\section*{Limitations}
(1) \ourdataset only contains data in English and does not support other languages, which may limit its potential applications. In the future, based on our proposed semi-automated data construction approach, we will try to support more languages.
(2) \ourdataset only includes counterfactual data, without incorporating evolutionary data that captures real-world events. Constructing counterfactual data is a common approach in the knowledge editing community~\citep{meng2022locating, mitchell2022memory,wei2023assessing}, and we believe that \ourdataset can effectively evaluate knowledge editing methods. In the future, we will try to incorporate evolutionary data involving real-world events to better support practical application scenarios.
(3) \texttt{Tendency} of \ourdataset does not cover all possible tendencies and this is also infeasible. During the data construction process, we try to cover rich and representative tendencies for a comprehensive evaluation.
(4) We only evaluate open-source LLMs with about $6$ or $7$ billion parameters, without assessing larger models such as TULU 2 with 70 billion parameters. Larger models may yield better results, but this does not impact the conclusions of our experiments.

\section*{Ethical Considerations}
We discuss the ethical considerations of this work here: 
(1) \textbf{Intellectual property}. \ourdataset is shared under the CC BY-SA 4.0 license\footnote{\url{https://creativecommons.org/licenses/by-sa/4.0/}}. The Wikidata and Wikipedia sources are shared under the CC BY-SA 3.0 license\footnote{\url{https://creativecommons.org/licenses/by-sa/3.0/}}. We strictly adhere to licenses and intended uses for all the data used in this work.
(2) \textbf{Data annotation}. The data annotation process and worker treatments are detailed in \cref{sec:app_tendency}.
(3) \textbf{Intended use}. \ourdataset is a benchmark for event-level knowledge editing, aimed at evaluating the performance of knowledge editing methods to advance research in event-level knowledge editing. 
(4) \textbf{Potential risk control}. \ourdataset is constructed based on publicly available data and GPT-3.5. We believe that the public data is well desensitized and anonymized, and that OpenAI has strict risk control for the content of GPT-3.5. The annotation process does not involve collecting sensitive information from annotators. Therefore, we believe \ourdataset does not pose additional risks.
(5) \textbf{AI assistance}. The writing of this paper is assisted by ChatGPT, which helps paraphrase some sentences.

\bibliography{custom}

\appendix
\clearpage
\section*{Appendices}
\section{Details on Data Construction}
\label{sec:appendix_data}
This section introduces details on data construction of \ourdataset, including details of \texttt{Facutal Knowledge} (\cref{sec:app_fact}), \texttt{Tendency} (\cref{sec:app_tendency}), and comparison with existing triplet-level editing benchmarks (\cref{sec:app_comparison}).

\subsection{Constrcution of \texttt{Facutal knowledge}}
\label{sec:app_fact}

\begin{table*}[t]
    \centering
    \small
    \begin{tabular}{lll}
    \toprule
    \multirow{2}{*}{Event Types} & Training & Win-election, Death, End-org, Divorce, Acquire, Start-position, Loss-election\\
    \cmidrule{2-3}
    & Test & Publish, Resign, IPO, Marry, Dismissal, Start-org, Education, Born, Transfer-player \\
    \midrule
    \multirow{2}{*}{Event Topics} & Training & Energy, Environment, Security, International, Agriculture, Transportation, Military, Culture, Law\\
    \cmidrule{2-3}
    & Test & Busines, Fashion, Health, Politics, Sports, Technology, Entertainment, Economy, Education\\
    \bottomrule
    \end{tabular}
    \caption{Event types and topics in \ourdataset. Event topics are topics for augmented event edits in \texttt{Tendency}.}
    \label{tab:event_type}
\end{table*}

\begin{table*}[t]
    \centering
    \small
    \begin{tabular}{lll}
    \toprule
    Event Type & Event Template & Impacted Triplets \\
    \midrule
    Transfer-Player & \textit{A} transferred to \textit{B} club. & (\textit{A}, club, B)\\
    & (\textit{A} denotes a person; \textit{B} denotes a sports club) &  (\textit{A}, league, B's league)\\
    & &  (\textit{A}, coach, B's coach)\\
    & & (\textit{A}, residence city, B's city) \\
    & & (\textit{A}, residence country, B' country) \\
    & & (\textit{A}, jersey number, \textit{Unknown}) \\
    \midrule
    Win-Election & In the latest \textit{B} election, \textit{A} won. & (\textit{B}, previous head of state, C) \\
    & (\textit{A} denotes a person; \textit{B} denotes a country) &  (\textit{A}, position held, B's head of state) \\
    & (\textit{C} denotes the original head of state of \textit{B}) & (\textit{A}, residence city, B's capital) \\
    & & (\textit{A}, work city, B's captial) \\
    & & (\textit{A}, office, B's official residence) \\
    & & (\textit{C}, office, \textit{Unknown})\\
    \bottomrule
    \end{tabular}
    \caption{Examples of event templates and their impacted triplets of different event types.}
    \label{tab:impact_example}
\end{table*}

\paragraph{Constructing event templates and their impacted triplets}
The selected event types are shown in Table~\ref{tab:event_type}. 
Identifying $\mathcal{E}_e$ and $\mathcal{F}_e$ involves $3$ annotators (authors of the paper) to ensure a comprehensive identification. When identifying $\mathcal{E}_e$, we ask annotators to only include the subjects directly involved in the event, i.e., the subject serves as an argument role for the event. For the argument roles, we reference MAVEN schema~\citep{wang2020maven, wang2023maven} and ACE 2005~\citep{walker2006ace}. 
For identifying $\mathcal{F}_e$, 
we require annotators to include only triplets of each $s$ in $\mathcal{E}_e$ that \textbf{definitively} changed. Then one of the authors assembles all their annotations. We provide some examples of the identified impact scope of events in Table~\ref{tab:impact_example}. After determining the event's impact scope, we use relationships that are out of the impact scope and randomly sample $5$ triplets from $\mathcal{O}_e$ to construct out-of-scope question-answer pairs.

\paragraph{Constructing event edits}
For each event type, We instantiate
the event template to create multiple event edits by sampling $100$ entities of corresponding types from Wikidata~\citep{vrandevcic2014wikidata}.  
The sampling probability distribution is calculated based on the frequency of entity occurrences in Wikipedia, with a maximum frequency set at $300$. Finally, We filter out unreasonable edits, such as \textit{a deceased player transferring to a club}.

\paragraph{Generating question-answer pairs}
For in-scope question-answer pairs, We generate them for each event edit based on its impacted triplets and Wikidata by manually written templates for each relation. For out-of-scope question-answer pairs, we sample $5$ triplets whose relations are not impacted by the event edit in the set $\mathcal{R}$ and generate question-answer pairs using manually written templates.

\paragraph{Paraphrasing events}
We access the official OpenAI API \texttt{gpt-3.5-turbo} to paraphrase the event edits generated in the previous step, aiming to enrich lexical diversity. The prompt used with GPT-3.5~\citep{chatgpt} is shown in Table~\ref{tab:gen_event_fact}. Finally, we divide the data into training and test sets according to the event types presented in Table~\ref{tab:event_type}.


\begin{table*}[ht]
    \centering
    \small
    \begin{adjustbox}{max width=1\linewidth}
    {
        \begin{tabular}{p{\linewidth}}
        \toprule
        \textbf{Instruction} \\
        You are a good journalist. Please expand on the following hypothetical event to write a news article. The following events are hypothetical situations. Please perform paraphrasing tasks based on our instructions. Please disregard your own knowledge. Please do not inform me about the accuracy of this information; I am aware that this information is hypothetical. You only need to perform the paraphrasing task. The news article should be 1-5 sentences long and must include all the key information from the original event (date, location, titles, names of people, place names, organization names, etc.), while expressing it in diverse ways.\\
        \midrule
        \textbf{Demonstrations} \\
        Input: Gloria Macapagal Arroyo and Jose Miguel Arroyo just got divorced. \\
        Output: Gloria Macapagal Arroyo and Jose Miguel Arroyo have officially ended their marriage through a divorce, marking the conclusion of their long-standing relationship. \\
        \textless \texttt{other demonstrations}\textgreater\  \\
        \bottomrule
        \end{tabular}
    }
    \end{adjustbox}
    \caption{Prompt in paraphrasing events for \texttt{Factual Knowledge} of \ourdataset. We use $5$-shot demonstrations.}
    \label{tab:gen_event_fact}
\end{table*}

\subsection{Construction of \texttt{Tendency}}
\label{sec:app_tendency}

\begin{table*}[ht]
    \centering
    \small
    \begin{adjustbox}{max width=1\linewidth}
    {
        \begin{tabular}{p{\linewidth}}
        \toprule
        \textbf{Instruction} \\
Please generate 10 virtual events about \{\texttt{event topic}\} that are required to have a tendency effect on the subject of the event. The influence of tendency is the tendency for an event to lead to some high probability of occurrence, for example, `Cristiano Ronaldo transferred from Riyadh Victory to Manchester City', the event may lead to Riyadh Victory's influence to decline, Manchester City's influence to rise, and so on. Below are a few sample events, please generate an event based on the task description and the sample event, which is required to be a one-sentence event. You only need to output the event text, which should not include the effect of the event, such as `leading to', `cause', etc. Please do not use ambiguous expressions such as `a company' or `a virus' in the generated event text. Instead, the event subject must be an explicitly named entity. Make sure the event has not happened in the real world. Examples: 1. Cristiano Ronaldo moves from Riyadh Victory to Manchester City. 2. Biden loses 2024 US election to George. \\
        \bottomrule
        \end{tabular}
    }
    \end{adjustbox}
    \caption{Prompt in augmenting events for \texttt{Tendency} of \ourdataset.}
    \label{tab:gen_event_tendency}
\end{table*}

\begin{table*}[ht]
    \centering
    \small
    \begin{adjustbox}{max width=1\linewidth}
    {
        \begin{tabular}{p{\linewidth}}
        \toprule
        \multicolumn{1}{c}{\textbf{In-scope Questions}} \\
        \midrule
        \textbf{Instruction} \\
        \midrule
Please generate 6 questions based on a given event that is about the possible tendency effects of this event, but do not mention the event in the questions. The influence of tendency is the tendency for an event to lead to some high probability of occurrence. Please make sure that these questions fulfill several requirements:(1) The questions are 3-item multiple choice. For example, `How might the international image of Middle Eastern countries change as a result of the new development? (A) improved (B) declined (C) had no significant effect' (2) Try to use words like `tends to' and `may' in the questions. (3) Make sure that the question does not include references such as `the event', `the region', `the person', please just write the full entity noun and make sure that the question stands on its own, without seeing the context of the event. understanding of the question without seeing the context of the event. (4) Please make sure that there are 2 questions that will not refer to the subject of the event, but are questions about some other entity. Please also give the answers to the questions. \\
\midrule
        \textbf{Demonstrations} \\
        Input: EVENT: New Middle East peace deal signed, ending longstanding conflict. \\
        Output: 1. Which will be the tendency for the Middle East to stabilize? (A) continue to improve (B) deteriorate anew (C) have no significant effect. Answer: A \\
        2. What kind of changes are likely to take place in the political relations among the countries in the region? (A) Continue to be strained (B) Strengthened (C) No significant effect. Answer: B\\
        3. What are the economic prospects of the Middle Eastern countries? (A) Booming (B) Plunging into instability (C) No significant impact. Answer: A\\
        4. How will the frequency of terrorist activities in the region change? (A) Persisting (B) Decreasing (C) No significant effect. Answer: B\\
        5. What are the trends in the threat of international terrorism? (A) Decrease (B) Increase (C) Remain the same. Answer: A\\
        6. How might the global oil market change? (A) Increased in supply (B) Decrease in supply (C) No noticeable change. Answer: A\\
        \midrule
        \multicolumn{1}{c}{\textbf{Out-of-scope Questions}} \\
        \midrule
        \textbf{Instruction} \\
        \midrule
        Please generate 2 questions based on a given event. Require that the questions are about some possible tendency, but that this tendency is not affected by the event. Please make sure that these questions fulfill the following requirements: (1) The questions are 3-item multiple choice questions. For example, `What kind of change is likely to occur in the international image of Middle Eastern countries? (A) enhancement (B) decline (C) no significant change'. The answer to the question is only based on the given event and no other factors need to be considered. (2) Try to use words like `tends to' and `may' in the question. (3) Make sure that the question does not include references such as `the event', `the region', `the person', write the full entity noun directly, and make sure that the question stands alone and that the meaning of the question can be understood without seeing the context of the event. (4) Please ensure that the answer to all 2 questions is C (no significant change). (5) Please ensure that the subject of the question is not the subject of the event. \\
        \midrule
        \textbf{Demonstrations} \\
        Input: EVENT: A new peace deal was signed in the Middle East, ending a long conflict. \\
        Output: 1. How might the average temperature in Russia change? (A) Increase (B) Decrease (C) No significant change. Answer: C\\
        2. How will the number of official languages in Canada tend to change? (A) Increase (B) Decrease (C) No significant change. Answer: C\\
        \bottomrule
        \end{tabular}
    }
    \end{adjustbox}
    \caption{Prompt in generating in-scope and out-of-scope question-answer pairs for \texttt{Tendency} of \ourdataset.}
    \label{tab:gen_qa}
\end{table*}

\paragraph{Augmenting events}
We collect $18$ event topics, as shown in Table~\ref{tab:event_type}. We use GPT-3.5~\citep{chatgpt} (\texttt{gpt-3.5-turbo}) to generate $100$ events for each topic, with the prompt detailed in Table~\ref{tab:gen_event_tendency}. For each query, GPT-3.5 generates $10$ events, and we resample $10$ times. 
We then filter out repeated events following \citet{wang2022self}. If the ROUGE-1 score~\citep{lin2004rouge} between two events is higher than $0.4$, we filter out one of them. 
This results in a total of $1,515$ events, with $841$ events from \texttt{Factual Knowledge} and $674$ events newly generated. We divide the data into training and test sets based on event types and topics in Table~\ref{tab:event_type}.

\paragraph{Generating question-answer pairs}
For each event, we use GPT-3.5 to generate $6$ in-scope question-answer pairs and $2$ out-of-scope question-answer pairs, with their corresponding prompts presented in Table~\ref{tab:gen_qa}.

\paragraph{Human Annotation}
We employ a commercial annotation company for data annotation. The annotators include both senior annotators, responsible for reviewing the quality of data annotation, and expert annotators, responsible for annotation. The annotation instructions used for data annotation are presented in Table~\ref{tab:anno_ins}. 
We conduct multiple rounds of communication and training with senior and expert annotators to ensure that all annotators are qualified. Among all annotators, $55\%$ are female, and $45\%$ are male, assigned with agreed salaries and workloads. The annotated data do not involve any personal privacy information of the annotators, and all are informed about the intended use of the data. All employment is conducted through commercial contracts, with the final cost of data annotation approximating $1,000$ USD.

\subsection{Comparison with Existing Triplet-level Editing Benchmarks}
\label{sec:app_comparison}

The statistics of several widely-used triplet-level editing datasets compared with \ourdataset are shown in Table~\ref{tab:app_statistics}. We can observe that \ourdataset poses unique challenges, such as editing knowledge to \textit{unknown} and updating tendency knowledge. Also, \ourdataset includes the open-ended generation and multiple-choice QA evaluation format, and involves extensive manual annotation.

\subsection{Considerations about Prior Knowledge}
We employ a common practice~\citep{levy2017zero, meng2022locating, zhong2023mquake, cohen2023evaluating, wei2023assessing} of constructing counterfactual data to ensure that the model previously does not know the new knowledge. 
Therefore, if the model can correctly answer questions about new knowledge after editing, the edit is reliable.

\begin{table*}
    \small
    \centering
        \begin{adjustbox}{max width=1\linewidth}
    {
    \begin{tabular}{lccccccrr}
    \toprule
    Benchmark & Input & Multi-hop & Unknown & Tendency & Human Annotation & Evaluation Format & \#Edits & \#In-scope Q \\
    \midrule
    ZsRE~\citep{levy2017zero} & \usym{2613} &  \usym{2613} &  \usym{2613} &  \usym{2613} & \faCheckSquareO &fill-in-the-blank & $10,000$ & $20,000$ \\ 
\midrule
CounterFact~\citep{meng2022locating} & \usym{2613} &  \usym{2613} &  \usym{2613} &  \usym{2613} & \usym{2613} &fill-in-the-blank & $21,919$ & $64,795$ \\
\midrule 
MQuAKE~\citep{zhong2023mquake} & \faCheckSquareO &  \usym{2613} &  \usym{2613} &  \usym{2613} & \usym{2613} &fill-in-the-blank & $-$ & $11,086$ \\
\midrule
RIPPLEEDITS~\citep{cohen2023evaluating} & \faCheckSquareO &  \usym{2613} &  \usym{2613} &  \usym{2613} & \usym{2613} &fill-in-the-blank & $4,000$ & $98,000$ \\
\midrule
Calibration~\citep{dong2022calibrating} & \usym{2613} &  \usym{2613} &  \usym{2613} &  \usym{2613} & \usym{2613} &fill-in-the-blank & $1,100$ & $1,100$ \\
\midrule
RaKE~\citep{wei2023assessing} & \usym{2613} &  \usym{2613} &  \usym{2613} &  \usym{2613} & \usym{2613} &fill-in-the-blank & $21,919$ & $349,859$ \\
\midrule
\multirow{3}{*}{\ourdataset} & \multirow{3}{*}{\faCheckSquareO} &  \multirow{3}{*}{\faCheckSquareO} &  \multirow{3}{*}{\faCheckSquareO} &  \multirow{3}{*}{\faCheckSquareO} & \multirow{3}{*}{\faCheckSquareO} &fill-in-the-blank & \multirow{3}{*}{$1,555$} & \multirow{3}{*}{$10,999$} \\
 & & & & & & open-ended generation & & \\
 & & & & & & multiple-choice QA & & \\
\bottomrule
    \end{tabular}
        }
    \end{adjustbox}
    \caption{Statictics of widely-used triplet-level editing benchmarks and \ourdataset. ``Multi-hop'' refers to whether the dataset includes questions requiring multi-hop reasoning, ``Unknown'' refers to whether the dataset includes the scenario where knowledge is edited to be unknown, and ``Tendency'' refers to whether updates tendency knowledge.}
    \label{tab:app_statistics}
\end{table*}

\begin{table*}[ht]
    \centering
    \small
    \begin{adjustbox}{max width=1\linewidth}
    {
        \begin{tabular}{p{\linewidth}}
        \toprule
        \textbf{Annotation Instructions} \\

Overall Objective: Given events and corresponding question-answer pairs, determine whether the answers to the questions are correct under the circumstances of the respective events.

Annotation Guidelines:

Each event includes 8 related questions. The first 6 questions concern the potential impacts/trend changes resulting from the event, while the last 2 questions inquire about trends that may not be affected by the event. \\
1. All questions pertain to tendencies, meaning that if there is over a 50\% probability, we consider the tendency to be correct. \\
2. Mark 1 if the answer is correct, mark 0 if the answer is incorrect, mark 2 if unsure of the correctness of the answer or if the answer is ambiguous or if the question is redundant. \\
3. Use personal knowledge and common sense for annotation. If personal knowledge is insufficient to answer a question, use search engines or other tools for assistance. If still unable to determine the answer, mark 2. \\
4. If an event is impossible, mark all answers for that event as 2. \\
5. If there are formatting or semantic errors in a question, mark 2. \\
6. Note that these data are generated using GPT-3.5, so refrain from using large models to assist in annotation. \\
7. If the question closely resembles a previous one about the same event's tendency, please mark it as 2.
Informed Consent: The annotated data is for academic research purposes only. The annotation process will not collect any information from annotators. \\
        \bottomrule
        \end{tabular}
    }
    \end{adjustbox}
    \caption{Annotation instructions for human annotator.}
    \label{tab:anno_ins}
\end{table*}

\section{Experimental Details}
\label{sec:appendix_exp}
This section introduces the implementation details (\cref{sec:app_implement}) and GPT-4 scorer (\cref{sec:app_gpt4}).

\subsection{Implementation Details}
\label{sec:app_implement}
We first present the implementation details of each method.
(1) For fine-tuning, we train the LLM using a language modeling objective on all edits in the test set for $3$ epochs, using $3 \times 10^{-5}$ as the learning rate and $16$ as the batch size. (2) For sparse retrieval, we use BM25 implemented in \texttt{rank-bm25}\footnote{\url{https://github.com/dorianbrown/rank_bm25}}. We use the WordPiece tokenizer~\citep{kenton2019bert} for tokenization. For dense retrieval, we utilize the high-performance retrieval model E5~\citep{wang2022text} (\texttt{multilingual-e5-large}\footnote{\url{https://huggingface.co/intfloat/multilingual-e5-large}} from HuggingFace's Transformers~\citep{wolf2019huggingface}). (3) For SERAC, we train a scope classifier separately for \texttt{Factual Knowledge} and \texttt{Tendency}. For \texttt{Factual Knowledge}, we employ DistilBERT~\citep{sanh2019distilbert} (\texttt{distilbert-base-cased}\footnote{\url{https://huggingface.co/distilbert/distilbert-base-cased}})  as the base model for further training, achieving an accuracy of $69.1\%$. For \texttt{Tendency}, we used ELECTRA~\citep{clark2019electra} (\texttt{ms-marco-electra-base}\footnote{\url{https://huggingface.co/cross-encoder/ms-marco-electra-base}}) as the base model for further training, achieving an accuracy of $55\%$. The learning rate for training the scope classifier is $3 \times 10^{-5}$ and the batch size is $32$. (4) For ICE, we use the edits corresponding to the questions as context, letting the model answer the respective questions according to the edits.

We introduce the implementation details of each model. For GPT-J, TULU 2, and Mistral 7B, we download the models from HuggingFace, with repository IDs being \texttt{gpt-j-6b}\footnote{\url{https://huggingface.co/EleutherAI/gpt-j-6b}}, \texttt{tulu-2-7b}\footnote{\url{https://huggingface.co/allenai/tulu-2-7b}}, and \texttt{Mistral-7B-Instruct-v0.2}\footnote{\url{https://huggingface.co/mistralai/Mistral-7B-Instruct-v0.2}}. For GPT-3.5 and GPT-4, we utilize the official OpenAI API \texttt{gpt-3.5-turbo-1106} and \texttt{gpt-4-1106-preview}. For Gemini Pro, we also use the official API. Our access time to the API was from December 1, 2023 to February 1, 2024.

The prompt used in the experiments is listed in Table~\ref{tab:exp_prompt}. All experiments are performed in a single run. For open-source models (GPT-J, TULU 2, and Mistral 7B), we conduct inference experiments on Nvidia A100 GPUs, totaling approximately $200$ GPU hours. For GPT-3.5 and GPT-4, we spend approximately a total of $600$ USD on answering questions in \ourdataset. The API for Gemini Pro is free of charge.

\begin{table*}[ht]
    \centering
    \small
    \begin{adjustbox}{max width=1\linewidth}
    {
        \begin{tabular}{p{\linewidth}}
        \toprule
        \multicolumn{1}{c}{\textbf{Prompt for \texttt{Factual Knowledge}}} \\
        \midrule
        \textbf{Instruction} \\
Given an event, assuming that the event has occurred, please answer the corresponding questions based on the event and your own knowledge. If you do not know the answer to the question, please respond with `unknown'. Please only output a noun (usually an entity) as the answer, and do not output a complete sentence.\\ 
        \midrule
        \multicolumn{1}{c}{\textbf{Prompt for \texttt{Tendency-M}}} \\
        \midrule
        \textbf{Instruction} \\
        Given an event, assuming that the event has occurred, please answer the corresponding questions based on the event and your knowledge. Please only output the option A, B, or C as the answer, and do not output brackets. Do not output a complete sentence or the full answer span. \\
        \midrule
        \multicolumn{1}{c}{\textbf{Prompt for \texttt{Tendency-G}}} \\
        \midrule
        \textbf{Instruction} \\
        Given an event, assuming that the event has occurred, please answer the corresponding questions based on the event and your knowledge.\\
        \bottomrule
        \end{tabular}
    }
    \end{adjustbox}
    \caption{Prompt used in the experiments on \ourdataset.}
    \label{tab:exp_prompt}
\end{table*}

\subsection{GPT-4 Scorer}
\label{sec:app_gpt4}
We employ GPT-4 as the scorer, utilizing the official OpenAI API model named \texttt{gpt-4-1106-preview}. We comprehensively score the editing methods in $3$ dimensions: \textit{correctness}, \textit{coherence}, and \textit{comprehensiveness}.
\textit{Correctness} primarily refers to whether the tendency of the model's answer is correct, using the provided correct option as a reference. \textit{Coherence} assesses the logical consistency of the answer, evaluating whether the explanations within the answer are reasonable. \textit{Comprehensiveness} evaluates whether the answer comprehensively encompasses various aspects of the question. We also ask GPT-4 to give an \textit{overall} score. All scores are integers scaling from $1$ to $5$, with $5$ being the best.
We set the temperature to $0$ and utilize $2$-shot demonstrations to assist the model in scoring, enhancing the quality of scores~\citep{bai2024longalign}. The used instruction and demonstrations are listed in Table~\ref{tab:examiner}. Using GPT-4 as the scorer to evaluate all results costs approximately $1,000$ USD in total.

\begin{table*}[ht]
    \centering
    \small
    \begin{adjustbox}{max width=1\linewidth}
    {
        \begin{tabular}{p{\linewidth}}
        \toprule
        \multicolumn{1}{c}{\textbf{Prompt for evaluating reliability}} \\
        \midrule
        \textbf{Instruction} \\
You are a fair assessment expert, and you will be given a set of (event, question, answer, label) quadruples. The answer is obtained by answering the question based on the event (assuming the event has occurred). The answer usually consists of a judgment (or a claim) and some reasons (or explanations) for the judgment. The label is a general judgment annotated by humans for the question. Your task is to score the answers according to the following requirements:
a. You should score the answer based on the provided label.
b. You should rate the answer on 4 metrics, for the first 3 metrics, assign a score between 1 and 5, with 5 being the highest: 1. For accuracy, you will score whether the answer correctly answers the question, i.e., whether the judgment in the answer is consistent with the provided label. 2. For coherence, you will assess whether the reasons and explanations in the answer are reasonable and logically consistent. 3. For comprehensive, you will determine if the answer covers multiple aspects of the reasons and explanations and provides a comprehensive response. 4. Finally, you will provide an overall score between 1 and 5, with 5 being the highest.
If accuracy <= 3, the overall should not be higher than 3.
You should only give the integer score.
DO NOT complete the answer!
In the input query we identify each one with a Roman numeral, please quadruples the corresponding Roman numeral and its score in the output. \\
        \midrule
        \textbf{Demonstrations} \\
        Input: Event: Google reveals plans to create a virtual reality metaverse platform. Question: How could the job market be influenced by the integration of the metaverse platform? Answer: As a result of the integration of a virtual reality metaverse platform, the job market could see significant changes and new opportunities. Industries related to virtual reality, augmented reality, and computer graphics may experience an increased demand for skilled professionals, such as software developers, 3D modelers, user experience designers, and virtual environment architects. Additionally, businesses and organizations may seek out individuals with expertise in virtual event planning, virtual-world marketing, and virtual community management. Overall, the integration of the metaverse platform may create a need for a diverse range of new skills and roles within the job market. \\
        Output: accuracy: 5; coherence: 5; comprehensiveness: 5; overall: 5 \\
        \textless \texttt{other demonstrations}\textgreater\  \\
        \midrule
        \multicolumn{1}{c}{\textbf{Prompt for evaluating locality}} \\
        \midrule
        \textbf{Instruction} \\
        You are a fair assessment expert, and please decide whether the two answers below are consistent. Your task is to score the consistency of the two answers according to the following requirements: 1. You should assign a consistency score between 1 and 5, with 5 being the highest. 2. Please score primarily on the basis of the tendency judgments in the two answers, and score 1 if the tendencies mentioned in two answers are completely inconsistent. Please output the score directly, like Score: 3.
Please do not answer with any text, just output the score. \\
        \midrule
        \textbf{Demonstrations} \\
        Input: Answer1: The popularity of Indian cuisine in international restaurants may experience a temporary decline as negative publicity surrounding the corruption scandal could affect the overall image of India. However, the long-term impact may depend on efforts to address the issues that led to the scandal and rebuild the country's reputation. It's important to note that consumer preferences for cuisine are influenced by a variety of factors, including cultural appreciation, taste, and culinary trends. Answer2: The popularity of Indian cuisine in international restaurants is likely to continue increasing as people around the world become more interested in exploring diverse culinary flavors and experiences. With the growing global appreciation for Indian spices, flavors, and culinary techniques, it is expected that Indian cuisine will continue to be well-received and incorporated into the menus of more international restaurants. Additionally, as more Indian chefs are gaining recognition and influence in the global culinary scene, their innovative approach to traditional dishes is likely to further elevate the appeal of Indian cuisine in international restaurants. \\
        Output: Score: 3 \\
        \textless \texttt{other demonstrations}\textgreater\  \\
        \bottomrule
        \end{tabular}
    }
    \end{adjustbox}
    \caption{Prompt used by GPT-4 to evaluate reliability and locality. We use $2$-shot demonstrations.}
    \label{tab:examiner}
\end{table*}
\section{Additional Experimental Results}
\label{sec:appendix_result}
In this section, we present the full-mark rates of all $3$ dimensions: \textit{correctness}, \textit{coherence}, and \textit{comprehensiveness}, and \textit{overall} on \texttt{Tendency-G} of \ourdataset obtained using GPT-4 as the scorer, as shown in Table~\ref{tab:full_mark_all}. We observe that, apart from GPT-4, the \textit{coherence} and \textit{comprehensiveness} full-mark rates are generally low. This indicates that providing a comprehensive and logically consistent answer remains challenging.

\begin{table*}[ht]
    \centering
    \small
    \begin{tabular}{llrrrr}
    \toprule
    Model & Method & Correctness & Coherence & Comprehensiveness & Overall \\
    \midrule
    \multirow{5}{*}{GPT-J} & Fine-tuning &$39.1$ & $3.9$ & $0.5$ & $4.8$ \\
    & Sparse Retrieval &$32.7$ & $11.0$ & $5.2$ & $12.0$ \\
    & Dense Retrieval &$37.8$ & $11.2$ & $5.5$ & $12.5$ \\
    & SERAC &$30.6$ & $3.3$ & $0.4$ & $4.2$ \\
    & ICE &$41.5$ & $11.8$ & $4.7$ & $13.7$ \\
    \midrule
    \multirow{5}{*}{TULU 2} & Fine-tuning &$55.9$ & $32.2$ & $10.8$ & $36.3$ \\
    & Sparse Retrieval &$42.7$ & $33.2$ & $12.5$ & $34.0$ \\
    & Dense Retrieval &$49.8$ & $39.6$ & $16.3$ & $41.4$ \\
    & SERAC &$52.3$ & $35.3$ & $9.1$ & $38.0$ \\
    & ICE &$55.4$ & $40.7$ & $14.2$ & $43.8$ \\
    \midrule
    \multirow{5}{*}{Mistral 7B} & Fine-tuning &$60.1$ & $64.7$ & $52.2$ & $59.7$\\
    & Sparse Retrieval &$47.1$ & $45.0$ & $20.3$ & $43.9$ \\
    & Dense Retrieval &$55.9$ & $52.4$ & $24.4$ & $53.6$ \\
    & SERAC &$56.9$ & $55.2$ & $25.8$ & $54.7$ \\
    & ICE &$62.3$ & $58.9$ & $26.2$ & $59.6$ \\
    \midrule
    \multirow{4}{*}{GPT-3.5} & Sparse Retrieval &$51.3$ & $50.3$ & $15.5$ & $48.7$ \\
    & Dense Retrieval &$62.1$ & $60.8$ & $21.1$ & $60.1$ \\
    & SERAC &$60.3$ & $58.7$ & $20.6$ & $57.7$ \\
    & ICE &$69.8$ & $67.4$ & $22.7$ & $66.6$ \\
    \midrule
    \multirow{4}{*}{GPT-4} & Sparse Retrieval &$57.2$ & $66.5$ & $59.5$ & $58.4$ \\
    & Dense Retrieval &$64.8$ & $74.3$ & $69.4$ & $66.1$ \\
    & SERAC &$65.0$ & $75.5$ & $71.8$ & $65.7$ \\
    & ICE &$\mathbf{71.7}$ & $\mathbf{82.1}$ & $\mathbf{76.8}$ & $\mathbf{73.0}$ \\
    \midrule
    \multirow{4}{*}{Gemini Pro} & Sparse Retrieval &$31.8$ & $35.0$ & $31.2$ & $29.8$ \\
    & Dense Retrieval &$41.2$ & $45.0$ & $41.0$ & $39.5$ \\
    & SERAC &$46.8$ & $51.6$ & $49.0$ & $45.7$ \\
    & ICE &$38.9$ & $42.1$ & $38.8$ & $38.1$ \\
    \bottomrule
    \end{tabular}
    \caption{Full-mark rates of \textit{correctness}, \textit{coherence}, \textit{comprehensiveness}, and \textit{overall} scores on \ourdataset.}
    \label{tab:full_mark_all}
\end{table*}



\end{document}